\documentclass[10pt]{article}

\usepackage[margin=0.85in]{geometry}
\usepackage[T1]{fontenc}
\usepackage{lmodern}
\usepackage{microtype}

\usepackage{amsmath,amssymb,amsthm,mathtools}
\usepackage{algorithm}
\usepackage{algpseudocode}
\usepackage{booktabs}
\usepackage{graphicx}
\usepackage[numbers,sort&compress]{natbib}
\usepackage[hidelinks]{hyperref}

\newtheorem{definition}{Definition}
\newtheorem{proposition}{Proposition}
\newtheorem{theorem}{Theorem}

\begin{document}

\title{Feature-Aware Edge Sampling for Approximate One-Step Laplacian Propagation}

\author{%
Jonathan Z. Sun\\
\small Charleston Southern University\\
\texttt{jsun@csuniv.edu}
\and
Zheng Wang\\
\small University of Miami\\
\small \texttt{zheng.wang@miami.edu}
\and
Kebin Xu\\
\small College of Charleston\\
\small \texttt{xuk@cofc.edu}
}
\date{}
\maketitle

\begin{abstract}
This paper studies how to approximate one-step combinatorial Laplacian messages in graph neural networks using edge sampling. We propose a spanning tree based hybrid sampler that combines feature-conditioned message scores and effective-resistance leverage scores, derive the expected one-step message error and obtain the optimal sampling probabilities and error frontier for a given expected edge budget. Preliminary experiments on synthetic community graphs show that message error decreases, while classification accuracy increases slightly, as the edge budget grows, and performance of the hybrid sampler is in between of optimal selection uniform random sampling. We leave multi-step propagation and evaluation on larger and real graphs for future work.
\end{abstract}

\vspace{0.75em}

\section{Introduction}

Graph neural networks have become a standard framework for learning from relational data. GCN, GraphSAGE, and GAT exemplify the message passing paradigm in which node representations are repeatedly updated by aggregating information from neighbors \cite{kipf2017gcn,hamilton2017graphsage,velickovic2018gat}. For a sparse propagation matrix $P$, a typical layer has the form
\begin{equation}
H^{(\ell+1)}
=\sigma_\ell\!\left(PH^{(\ell)}W_\ell\right),
\qquad H^{(0)}=X,
\label{eq:gnnlayer}
\end{equation}
where $X$ is the initial node feature matrix, $P$ is the graph propagation matrix, and $H^{\ell}$, $W^{\ell}$, $\sigma_\ell$ are node representation matrix, learnable weight matrix, and activation function at layer $\ell$. 

\subsection{Motivation}
Although the cost of one sparse propagation is linear in the number of retained edges, repeated aggregation on very large graphs is expensive in time and memory. This scalability problem has motivated a variety of sampling schemes such as neighborhood sampling, layer-wise importance sampling, subgraph sampling, clustering, graph reduction, and model simplification \cite{hamilton2017graphsage,chen2018fastgcn,huang2018adaptive,zou2019ladies,chiang2019clustergcn,zeng2020graphsaint,wu2019sgc,hashemi2024survey}.

Graph sparsification replaces a graph by a substantially smaller weighted subgraph while preserving selected properties of the original graph. Classical cut and spectral sparsifiers provide strong guarantees for cuts, Laplacian quadratic forms, and related structural quantities \cite{eppstein1997sparsification,karger1999random,spielman2011spectral,spielman2011effective,batson2012twice}. However, these guarantees are task agnostic. In general, for a graph Laplacian $L$, a spectral sparsifier $L_H$ seeks to satisfy
\begin{equation}
(1-\epsilon)x^\top Lx
\le x^\top L_Hx
\le(1+\epsilon)x^\top Lx
\quad\text{for every }x\in\mathbb{R}^{n}.
\label{eq:spectral}
\end{equation}

However, a graph neural network (GNN) does not ordinarily explore all directions in $\mathbb{R}^{n}$. 
Instead of querying the graph through arbitrary vectors, a GNN repeatedly propagates a structured and evolving collection of node features \cite{kipf2017gcn,hamilton2017graphsage,velickovic2018gat}. 
Starting from a feature matrix $X\in\mathbb{R}^{n\times d}$, its linear propagation stages repeatedly generate signals related to $X,PX,P^2X,\ldots, P^nX$. Moreover, analysis of simplified GCN already illustrates the close relationship between graph propagation and polynomial graph filtering \cite{wu2019sgc}; while recent work on spectral GNN further reveals the relationship more explicitly \cite{ding2025spectral}.

This motivates our main problem of study:
\emph{How to design heuristic sparsifiers for GNN to preserve most feature signals used for message passing without paying
the cost of preserving the original graph on every spectral direction?}

\subsection{Idea}

In this paper, we study the above problem for one-step combinatorial Laplacian propagation on fixed input features. The idea is to retain a spanning tree and then independently sample the remaining edges, combining feature-conditioned message scores with effective-resistance scores. We derive the exact expected squared message error, the minimum error for each expected edge budget, and the hybrid proposal's excess over that minimum. In a controlled synthetic community graph experiment (with preset graph structures and feature conditions), message error decreases as the edge budget grows, while node classification accuracy remains similar across the tested sampling rules. Schemes and experiments for standard multi-layer GNNs and real world graphs remain in future work. 

Recent work already demonstrated the potential of GNN sparsification. FastGAT applies effective-resistance-based spectral sparsification to attention networks \cite{srinivasa2020fastgat, shen2024gper, chen2024demystifying, zhang2025mog, das2025sgsgnn, ding2025spectral}. Therefore we are not making ``yet another edge score'', but instead, interested in designing a \emph{message-preserving sparsifier}. 

We study how the graph Laplacian \(L\) acts on the feature matrix \(X\) and consider message preservation, i.e., the propagation result on the sparsified graph should be close to that on the original graph. Our goal is to construct a reusable sparse graph of given edge budget (amount of edges retained) as a preprocessing before downstream model training. Our method retains a spanning tree and independently samples the remaining edges, and reweights each sampled edge by the inverse of its inclusion probability. It then derives the expected one-step message error and obtains the optimal sampling probabilities and error frontier for a given expected edge budget. By combining a spanning tree backbone with the proposed feature-conditioned message score and the classic effective-resistance leverage score, we provide the algorithm of a hybrid sampler and quantify its error relative to optimal message allocation. 

\subsection{Contributions}

The goal of sampling here is computational cost in time and memory: to reduce the number of edges used in repeated propagation, while limiting a specified approximation error and not sacrificing predictive accuracy. Here ``message approximation'' refers to approximating the one-step action $LX$ for the fixed input feature matrix $X$ with $L_HX$. 

Major contributions in this paper are as follows. 

\begin{enumerate}
\item We define one-step message approximation through the error in applying combinatorial Laplacian to the input features. For independent edge sampling, we derive an unbiased estimator, its exact expected squared error, and the budget dependent optimal allocation with a fixed spanning tree backbone.
\item We design a hybrid scheme that considers both feature-conditioned message importance and effective-resistance importance and also retains a spanning-tree backbone to guarantee sampled graph's connectivity, and give exact decomposition of its message-error excess relative to the optimal message allocation. 

\item We perform preliminary experiments to support the soundness of our algorithm and analysis. Experimental results meet expectation, although not significant due to the quantity and quality of synthetic data we generated. 

\item We also present the feature-generated Krylov space as a direction for future work on multi-step propagation. 
\end{enumerate}

\subsection{Paper Organization}

The paper is organized as follows. Section~\ref{sec:related} reviews three lines of related work: classical graph sparsification as background, GNN computation sampling as a strategy of saving training time and memory, and GNN-oriented graph reduction as the closest context of reference. Section~\ref{sec:problem} introduces the terminologies and formulations to define a weighted graph and its Laplacian, edge message score, and approximation objectives; it also introduces Krylov-space preservation as foundation for future analysis. Section~\ref{sec:method} specifies the tree-based hybrid sampler, derives its expected one-step message error and budget-optimal allocation, and quantifies the hybrid algorithm's excess error. Section~\ref{sec:experiments} performs a controlled experiments with preset conditions on modest size synthetic data, which shows message error decreases with the edge budget grows, while node classification accuracy remains similar with no loss. Section~\ref{sec:conclusion} summarizes these preliminary findings and proposes future work for multi-layer propagation models and real graphs.

\section{Related Work}
\label{sec:related}

In this section we summarize the following three lines of work as they represent the methodological background, the complementary strategy for computational efficiency, and the closest competing approaches to reference to. 

\subsection{Classical Graph Sparsification}

Graph sparsification provides combinatorial algorithmic approaches to tackling the problem. Instead of sampling a temporary computation graph during a training iteration, it constructs a persistent sparse graph that approximates selected properties of the original graph. 

Eppstein et al. introduced algorithmic sparsification using sparse certificates as a technique for accelerating dynamic graph algorithms \cite{eppstein1997sparsification}. 
Karger developed random edge sampling methods that preserve graph cuts with high probability and used sparse sampled skeletons in cut, flow, and network-design algorithms \cite{karger1999random}. 
Spielman and Teng developed spectral sparsification to strengthened the preservation objective to Laplacian quadratic forms, resulting in nearly-linear-time graph algorithms \cite{spielman2011spectral}. Spielman and Srivastava then showed that sampling edges according to effective resistance yields high-quality spectral sparsifiers with $O(n\log n/\epsilon^2)$ edges \cite{spielman2011effective}. 
Batson, Spielman, and Srivastava subsequently presented twice-Ramanujan sparsifier to prove the existence of spectral sparsifiers with number of edges linear in $n$ \cite{batson2012twice}. 
Fung et al. presented a broader framework that encloses multiple sparsification measures, including edge strength, connectivity, and effective resistance \cite{fung2019framework}.

The objective of these classical combinatorial sparsifiers is to support arbitrary graph algorithms independent of any particular downstream machine learning models. However, ideas and results can be borrowed to inspire heuristics, particularly when the downstream model works primarily on low dimensional node signals.

\subsection{Sampling for Scalable GNN Training}

This line of work direct to reducing the cost of GNN training by sampling the computation graph. GraphSAGE samples and aggregates local neighborhoods to enable inductive representation learning \cite{hamilton2017graphsage}. FastGCN interprets graph convolution as an integral transform and applies importance sampling to obtain efficient mini-batch estimators \cite{chen2018fastgcn}. Adaptive sampling methods explicitly seek variance reduction \cite{huang2018adaptive}. LADIES uses layer-dependent importance sampling conditioned on nodes sampled in the upper layer \cite{zou2019ladies}. Cluster-GCN exploits graph clustering to train on densely connected blocks \cite{chiang2019clustergcn}. GraphSAINT samples subgraphs and applies normalization to reduce sampling bias \cite{zeng2020graphsaint}. DropEdge randomly removes edges during training and was proposed primarily as a regularization mechanism against overfitting and over-smoothing \cite{rong2020dropedge}.

These methods are relevant to our work because they demonstrate that sampling distributions can be designed to reduce estimator variance and training cost. However, our approach is  different: our proposed method produces a \emph{single persistent sparse graph} as a preprocessing, so its cost can be amortized across epochs and iterations using the same graph. Reusing the sparsified graph across architectures or downstream tasks is possible only when the propagation requests are compatible with the feature-conditioned sampling target; it is not guaranteed by the one-step analysis. Different from this line of work, our focus is the graph-level preservation objective and its feature-generated subspace interpretation.

\subsection{Graph Reduction and GNN-Oriented Sparsification}

Graph reduction includes sparsification, coarsening, condensation, and related summarization techniques \cite{hashemi2024survey}. Empirical evaluations demonstrated that preservation of conventional graph metrics does not automatically imply preservation of GNN behavior \cite{chen2024demystifying}. Vatter et al.\ studied graph sparsification jointly with GNN sampling and showed that substantial graph reduction can coexist with competitive learning performance \cite{vatter2024size}.The following methods were particularly designed for sparsification with GNNs. 

FastGAT uses effective-resistance-based spectral sparsification to reduce the cost of graph attention networks while providing a theoretical connection between spectral approximation and learned features \cite{srinivasa2020fastgat}. GPER uses effective resistance to identify edges for rewiring and preprocessing, targeting over-smoothing and over-squashing \cite{shen2024gper}. Spectral Greedy Graph Coresets select representative ego-graphs using spectral embeddings, addressing a related but node/ego-graph selection problem rather than persistent edge sparsification \cite{ding2024coresets}. Mixture-of-Graphs dynamically selects locally appropriate pruning experts and mixes sparse graphs \cite{zhang2025mog}. SGS-GNN learns supervised edge-sampling distributions and uses label-related objectives to improve sparse-graph prediction \cite{das2025sgsgnn}. Ding et al.\ use Laplacian sparsification to approximate polynomial filters in large-scale spectral GNNs and provide corresponding theoretical guarantees \cite{ding2025spectral}.

Comparing to the above methods, the scheme proposed in the next is unique in 1) its message score is derived from the one-step action of the combinatorial Laplacian on the input features rather than learned from labels or a downstream supervised loss; 2) it retains a spanning tree backbone and also take into consideration effective-resistance to yield a hybrid algorithm; and 3) its analysis concerns message error on the fixed feature matrix, while also consider Krylov-space preservation as a future direction to study multi-layer propagation. 

\section{Terminologies and Problem Formulation}
\label{sec:problem}

Let $G=(V,E,w,X)$ be a connected, undirected, weighted attributed graph with $n=|V|$, $m=|E|$, positive edge weights $w_e$, and feature matrix $X\in\mathbb{R}^{n\times d}$. Let $B\in\mathbb{R}^{m\times n}$ be an oriented incidence matrix and $W=\operatorname{diag}(w_e)$. The combinatorial Laplacian is
\begin{equation}
L=B^\top WB=\sum_{e\in E}w_e b_eb_e^\top,
\end{equation}
where $b_e^\top$ is the incidence row associated with edge $e$.

Applying $L$ to $X$ gives
\begin{equation}
LX=\sum_{e\in E}M_e(X),
\qquad
M_e(X)=w_e b_eb_e^\top X.
\label{eq:edge-message}
\end{equation}
For $e=(u,v)$,
\begin{equation}
\|M_e(X)\|_F
=\|b_e\|_2\,w_e\|b_e^\top X\|_2
=\sqrt{2}\,w_e\|x_u-x_v\|_2.
\label{eq:message-score}
\end{equation}
Here $M_e(X)$ the contribution of edge $e$ to the matrix $LX$. That is, an edge has high one-step message importance when it carries a large feature difference under Laplacian propagation.

\begin{definition}[Message-preserving sparsifier]
For feature matrix $X$, budget $q$, and tolerance $\epsilon>0$, a weighted subgraph $H=(V,E_H,\widetilde w)$ with $|E_H|\le q$ is an $(X,\epsilon)$ message-preserving sparsifier if
\begin{equation}
\|LX-L_HX\|_F\le\epsilon\|LX\|_F.
\end{equation}
\end{definition}

Note that this definition is weaker than the full spectral sparsification in Eq.~\eqref{eq:spectral}. It asks for accurate action only on the columns of $X$, not on every vector in $\mathbb{R}^n$.

\subsection{Subspace and Krylov Preservation}

Let $\mathcal{S}\subseteq\mathbb{R}^n$ be a subspace of $\mathbb{R}^n$.

\begin{definition}[Subspace spectral sparsifier]
$H$ is an $(\mathcal{S},\epsilon)$ subspace spectral sparsifier of $G$ if
\begin{equation}
(1-\epsilon)x^\top Lx
\le x^\top L_Hx
\le(1+\epsilon)x^\top Lx
\end{equation}
for every $x\in\mathcal{S}$.
\end{definition}
When $\mathcal{S}=\mathbb{R}^n$, this becomes the classical definition of the spectral sparsification. For attributed graphs, the straightforward choice is let $\mathcal{S}=\operatorname{span}(X)$. Therefore for a $k$-step propagation process, we propose the feature-generated block Krylov space denoted by
\begin{equation}
\mathcal{K}_k(P,X)
=\operatorname{span}\{X,PX,P^2X,\ldots,P^kX\}.
\label{eq:krylov}
\end{equation}
This construction is motivated by the observation that polynomial graph filters and linear message-passing models act inside spaces generated by repeated applications of the graph operator \cite{wu2019sgc,ding2025spectral}. The effective dimension of $\mathcal{K}_k(P,X)$ could be much smaller than $n$, particularly when $d\ll n$ and the propagated signals are highly correlated.

Therefore the intrinsic question behind sparsification is sharper than a GNN stability statement. The dimension of the span alone may not determine the required sample size; the geometry and concentration of the edge-message vectors within that span matters. 

A valid guarantee would address whether the operator is the combinatorial Laplacian or a normalized propagation matrix, whether the target is a quadratic form or the operator action on feature columns, and whether the result holds in expectation or with high probability. These distinctions of sampling objectives would lead to different heuristics. The general problem, for which the current paper addresses a first step, is as follows. 

\emph{How many sampled edges are sufficient to approximate $L$ or $P$ on $\mathcal{K}_k(P,X)$, and under what conditions can this amount be substantially smaller than the edge budget needed for a full spectral sparsifier?}

\section{Tree-Based Feature-Aware Sampling and Its Analysis}
\label{sec:method}

This section presents the proposed main method: a hybrid sampling algorithm for one-step propagation. Given a graph and its node-feature matrix, we keep a spanning tree and sample the remaining edges with the goal of approximating the Laplacian message $LX$ using fewer edges. We first show that inverse-probability reweighting gives an unbiased estimate and derive its exact expected squared error. We then use that error to characterize the best allocation of a fixed sampling budget and assess a hybrid proposal that, in addition to message importance, also takes effective resistance into consideration. 

\subsection{Unbiased Message Sampling}

Let $T$ be a fixed spanning tree and $J=E\setminus T$ be the remaining optional edges excluding the tree. The tree edges are retained with their original weights; for $e\in J$, let $\xi_e\sim\mathrm{Bernoulli}(p_e)$ be independent and assigned weight $w_e/p_e$ when retained. Denote
\[
M_e(X)=w_eb_eb_e^\top X,\qquad s_e=\|M_e(X)\|_F.
\]
Then we have
\begin{equation}
L_HX-LX=\sum_{e\in J}\left(\frac{\xi_e}{p_e}-1\right)M_e(X).
\label{eq:message-error-decomp}
\end{equation}
Since $\mathbb{E}\left[
\xi_e\right] = p_e$, 
each summand in Equation~\eqref{eq:message-error-decomp} is centered and therefore $\mathbb{E}[L_HX]=LX$. Since the summands are centered and independent, the cross inner products are zero in expectation,  yielding the exact identity
\begin{equation}
V(p):=\mathbb{E}\|L_HX-LX\|_F^2
=\sum_{e\in J}s_e^2\left(\frac{1}{p_e}-1\right).
\label{eq:variance}
\end{equation}
That is, the deterministic spanning tree contributes no sampling variance; its cost is the $n-1$ edges removed from the randomized budget. Equation~\eqref{eq:variance} is not a spectral or prediction guarantee but the exact message error objective. 

\subsection{Exact Budget Error Frontier}

Assume $s_e>0$ for the optional edges and fix an expected optional edge budget $B$ with $0<B<|J|$. The constrained problem is
\begin{equation}
V^*(B)=\min_{\substack{0<p_e\le1\\\sum_{e\in J}p_e=B}}V(p).
\label{eq:opt}
\end{equation}
The next theorem gives the minimum error for every budget, including the transition points at which high score edges are selected with probability 1.  

\begin{theorem}[Fixed-backbone message-error frontier]
\label{th:main}
There is a unique $\lambda>0$ such that
\begin{equation}
p_e^*(B)=\min\left\{1,\frac{s_e}{\lambda}\right\},
\qquad \sum_{e\in J}p_e^*(B)=B.
\label{eq:pstar}
\end{equation}
Let $\mathcal{A}=\{e:s_e\ge\lambda\}$ be the saturated edges and $\mathcal{F}=J\setminus\mathcal{A}$. Then
\begin{equation}
V^*(B)=\lambda\sum_{e\in\mathcal{F}}s_e-\sum_{e\in\mathcal{F}}s_e^2,
\qquad
\frac{dV^*(B)}{dB}=-\lambda^2
\label{eq:frontier}
\end{equation}
within each interval on which $\mathcal{A}$ is fixed. In particular, the marginal decrease in minimum expected squared error per additional expected edge is $\lambda^2$; the frontier is continuous and piecewise smooth, with breakpoints when an edge enters or leaves $\mathcal{A}$.
\end{theorem}

\begin{proof}
For each $e\in J$, the function
$f_e(p_e)=s_e^2(1/p_e-1)$ is strictly convex on $(0,1]$, since
$f_e''(p_e)=2s_e^2/p_e^3>0$. Thus, the objective is strictly convex,
and any feasible point satisfying the Karush-Kuhn-Tucker (KKT) conditions~\cite[Sec.~5.5.3]{boyd2004convex} is the unique
global minimizer.

Introduce a multiplier $\eta$ for the equality constraint
$\sum_{e\in J}p_e=B$ and multipliers $\mu_e\ge 0$ for the upper
constraints $p_e\le 1$. The Lagrangian is
\[
\mathcal{L}(p,\eta,\mu)
=
\sum_{e\in J}s_e^2\left(\frac{1}{p_e}-1\right)
+\eta\left(\sum_{e\in J}p_e-B\right)
+\sum_{e\in J}\mu_e(p_e-1).
\]
The stationarity and complementary-slackness conditions are
\[
-\frac{s_e^2}{p_e^2}+\eta+\mu_e=0,
\qquad
\mu_e(p_e-1)=0,
\qquad
\mu_e\ge 0.
\]
Because $B<|J|$, at least one edge has $p_e<1$. For such an edge,
complementary slackness gives $\mu_e=0$, and stationarity yields
$\eta=s_e^2/p_e^2>0$.
If $p_e<1$, then $\mu_e=0$, hence
$p_e=s_e/\sqrt{\eta}$. 
Otherwise if $p_e=1$, stationarity gives
$\mu_e=s_e^2-\eta\ge 0$, hence $s_e\ge\sqrt{\eta}$.
Let $\lambda=\sqrt{\eta}$ and combine these two cases as
\[
p_e=\min\left\{1,\frac{s_e}{\lambda}\right\}.
\]
The budget constraint then requires
\[
\sum_{e\in J}\min\left\{1,\frac{s_e}{\lambda}\right\}=B.
\]
The left-hand side is continuous in $\lambda$, decreases from $|J|$
to $0$, and is strictly decreasing whenever its value is below
$|J|$. Since $0<B<|J|$, this equation has a unique solution
$\lambda>0$. This proves Eq.~\eqref{eq:pstar} and, by strict
convexity, the optimal solution is unique.

Let $\mathcal{A}=\{e\in J:s_e\ge\lambda\}$ and
$\mathcal{F}=J\setminus\mathcal{A}$. For $e\in\mathcal{F}$,
$p_e^*=s_e/\lambda$, so its contribution to $V^*(B)$ is
\[
s_e^2\left(\frac{1}{p_e^*}-1\right)
=\lambda s_e-s_e^2.
\]
For $e\in\mathcal{A}$, $p_e^*=1$, so its contribution is zero.
Summing over $\mathcal{F}$ gives the first identity in
Eq.~\eqref{eq:frontier}.

When $\mathcal{A}$ is fixed, let
$S_{\mathcal{F}}=\sum_{e\in\mathcal{F}}s_e$. The budget equation gives
\[
B=|\mathcal{A}|+\frac{S_{\mathcal{F}}}{\lambda},
\qquad
\lambda=\frac{S_{\mathcal{F}}}{B-|\mathcal{A}|}.
\]
Consequently,
\[
V^*(B)
=
\frac{S_{\mathcal{F}}^2}{B-|\mathcal{A}|}
-\sum_{e\in\mathcal{F}}s_e^2,
\]
and differentiating within an interval on which $\mathcal{A}$ is
fixed yields
\[
\frac{dV^*(B)}{dB}
=
-\frac{S_{\mathcal{F}}^2}{(B-|\mathcal{A}|)^2}
=-\lambda^2.
\]
At a breakpoint where an edge becomes saturated, the formulas for $V^*(B)$ on the two adjacent budget intervals agree because that edge contributes zero error at $p_e^*=1$.
\end{proof}

This theorem characterizes the solution to a convex program. Its role here is to provide the exact benchmark for a message estimator that reserves a prescribed connectivity backbone. The proof also shows why the uncapped proportional-allocation rule alone is insufficient: the variance curve changes whenever an edge reaches an inclusion probability of one.

\subsection{Structural- and Message-Preserving Sampling}

The message-optimal allocation in Eq.~\eqref{eq:pstar} uses only the
feature-conditioned scores $s_e$. This subsection defines a feasible
proposal that also uses effective-resistance leverage scores $\tau_e$,
which encode structural information from the graph Laplacian. The goal
is to compare the two allocations at the same expected edge budget and
to quantify the one-step message-error cost of incorporating
structural scores, which is given in the following proposition. 

For $e\in J$, let $s_e$ be the feature-conditioned message score from
Eq.~\eqref{eq:message-score} and let $\tau_e=w_eb_e^\top L^\dagger b_e$ be the effective-resistance
leverage score~\cite{spielman2011effective}.
Normalize the two score families:
\begin{equation}
r_e=\frac{\tau_e}{\sum_{f\in J}\tau_f},\qquad
m_e=\frac{s_e}{\sum_{f\in J}s_f},\qquad
\pi_e(\alpha)=\alpha r_e+(1-\alpha)m_e,
\label{eq:mixture}
\end{equation}
where $0\le\alpha\le1$. For a target expected optional-edge budget
$B$, choose $c_\alpha$ so that
$p_e^\alpha=\min\{1,c_\alpha\pi_e(\alpha)\}$ satisfies
$\sum_{e\in J}p_e^\alpha=B$. This mixture is a feasible proposal, not
the message-optimal allocation in Eq.~\eqref{eq:pstar}.

\begin{proposition}[Exact excess error of a feasible proposal]
Let $p$ be any feasible inclusion-probability vector with
$0<p_e\le1$ and $\sum_{e\in J}p_e=B$, and let $p^*$, $\lambda$,
$\mathcal{A}$, and $\mathcal{F}$ be as defined above. Then
\begin{equation}
V(p)-V^*(B) \;\ge0.
\label{eq:excess-error}
\end{equation}
\end{proposition}

\begin{proof} We will establish the following equation: 
\begin{equation}
V(p)-V^*(B)
=\sum_{e\in J}
\frac{s_e^2(p_e-p_e^*)^2}{p_e(p_e^*)^2}
+\sum_{e\in\mathcal{A}}(s_e^2-\lambda^2)(1-p_e).
\label{eq:excess-error-sum}
\end{equation}

For each edge, we know the following algebraic identity holds:
\begin{equation}
\frac{s_e^2}{p_e}-\frac{s_e^2}{p_e^*}
=\frac{s_e^2(p_e-p_e^*)^2}{p_e(p_e^*)^2} 
-\frac{s_e^2}{(p_e^*)^2}(p_e-p_e^*).
\label{eq:algebraic-identity}
\end{equation}
Therefore we only need to show that summing the 2nd term on RHS of ~\eqref{eq:algebraic-identity} over all edges results in the 2nd term on RHS of~\eqref{eq:excess-error-sum}. For $e\in\mathcal{F}$,
$s_e^2/(p_e^*)^2=\lambda^2$, whereas for $e\in\mathcal{A}$,
$p_e^*=1$. Thus we sum the 2nd term on  RHS of ~\eqref{eq:algebraic-identity} for all edges and get
\begin{equation}
-\lambda^2\sum_{e\in\mathcal{F}}(p_e-p_e^*)
-\sum_{e\in\mathcal{A}}s_e^2(p_e-1)
\label{eq:linear-terms}
\end{equation}
Because $p$ and $p^*$ have the same budget, we have 
\begin{equation}
\sum_{e\in\mathcal{F}}(p_e-p_e^*)
=\sum_{e\in\mathcal{A}}(1-p_e)
\label{eq:budget-identity}
\end{equation}
Substituting ~\eqref{eq:budget-identity} into the first term of
~\eqref{eq:linear-terms} gives
\[
\sum_{e\in\mathcal{A}}(s_e^2-\lambda^2)(1-p_e),
\]
establishing Eq.~\eqref{eq:excess-error-sum}. It's obvious that each term on RHS of~\eqref{eq:excess-error-sum} is non-negative. The position is proved. 
\end{proof}

Eq.~\eqref{eq:excess-error} applies directly to the calibrated hybrid probabilities $p^\alpha$ and quantifies their message error cost relative to the best allocation at the same expected edge count. The proposition is a comparison identity that separates the proposal's excess message error into two
costs: the deviation from the message-optimal probabilities, and the
undersampling edges that the message-optimal rule would retain with
probability one. Applied to the calibrated probabilities
$p^\alpha$, it lets the algorithm and experiments compare the
structure- and message-preserving proposal with the message-only benchmark at a matched budget.

\subsection{Sampling Algorithm}

\begin{algorithm}[t]
\caption{Tree-based hybrid message sampling}
\label{alg:smps}
\begin{algorithmic}[1]
\Require Connected weighted graph $G=(V,E,w)$, feature matrix $X$, total expected edge budget $Q\ge n-1$, mixture parameter $\alpha$
\Ensure Sparse weighted graph $H$
\State Construct a spanning tree $T$ and retain its edges at original weights
\State For each $e\in E\setminus T$, compute $s_e=\|w_eb_eb_e^\top X\|_F$ and $\tau_e=w_eb_e^\top L^\dagger b_e$
\State Normalize the scores and form $\pi_e(\alpha)$ using Eq.~\eqref{eq:mixture}
\State Set $p_e=\min\{1,c_\alpha\pi_e(\alpha)\}$ with $\sum_{e\notin T}p_e=Q-(n-1)$
\State Independently sample each optional edge; reweight each retained edge by $1/p_e$
\State \Return $H$ containing $T$ and the sampled optional edges
\end{algorithmic}
\end{algorithm}

The sampling algorithm is given in Algorithm~\ref{alg:smps}. The spanning tree backbone alone gives no cut, spectral, message, or prediction guarantee, but only provides connectivity. The algorithm then computes scores for the remaining edges, calibrates their
inclusion probabilities to meet the expected edge budget, and samples
and reweights those edges. Computing the feature-conditioned message scores costs $O(md)$ arithmetic. Once the mixture probabilities are calibrated, forming
them and sampling the optional edges costs $O(m)$; the cost of
calibration depends on the implementation. Exact effective-resistance
scores can be expensive to compute, motivating approximate scores for
large graphs \cite{spielman2011effective}. If the sampled graph is
reused for propagation, each subsequent step costs $O(Qd)$ in
expectation, compared with $O(md)$ on the original graph. 

The analysis only applies to the one-step action of the combinatorial
Laplacian on the fixed input feature matrix $X$. It does not establish
guarantees for normalized propagation matrices, nonlinear hidden
representations, multi-layer networks, or feature-generated Krylov
spaces. Extending the analysis to these more realistic circumstances is left for future work.

\section{Experiments}
\label{sec:experiments}

The small synthetic pilot in this section only evaluates one-step message reconstruction and a simple downstream prediction probe. Purpose of the pilot is to find supportive evidences for the soundness of our heuristic sampling scheme and mathematical analysis. 

\subsection{Setup}

Ten connected undirected stochastic block model graphs were generated with seeds 20261106--20261115. Each graph has 120 vertices in three balanced classes, with within-class and between-class edge probabilities 0.18 and 0.02, respectively; disconnected draws were rejected. Unit edge weights were used. Each node has 12 features: four class-specific coordinates have mean 0.7, and independent Gaussian noise with standard deviation 1.2 is added.

Each graph uses a randomized minimum spanning tree shared by all samplers. For the non-tree edges, we compare the uniform random sampling, message score only sampling, and 50/50 effective resistance and message score sampling. We set target edge ratios (budgets) to be 30\%, 40\%, \ldots, 90\%, 95\%, and 99\%. The inclusion probabilities are calibrated to the expected total edge budget, and sampled edges are reweighted by their inverse probabilities. We perform each test on 10 graph instances, and run five independent sampling draws per graph, method, and budget. Each reported data point is the average with standard deviation. The random seed is $10^6s+10^4j+r_{\mathrm{code}}+17h$, where $s$ is the graph seed, $j$ the draw index, $r_{\mathrm{code}}$ the ratio in thousandths, and $h$ the method index. 

We measure relative message error $\|L_HX-LX\|_F/\|LX\|_F$. For prediction, we apply one residual step, $X'=X-\eta L_HX$, with $\eta=0.5/\lambda_{\max}(L)$, then train logistic regression ($C=1$) on the first 20 vertices per class and test on the remaining. Finally, this linear probe is very preliminary and not comparable to a standard GCN, GraphSAGE, or GAT.

\subsection{Results}

\begin{figure}[t]
\centering
\includegraphics[width=0.45\textwidth]{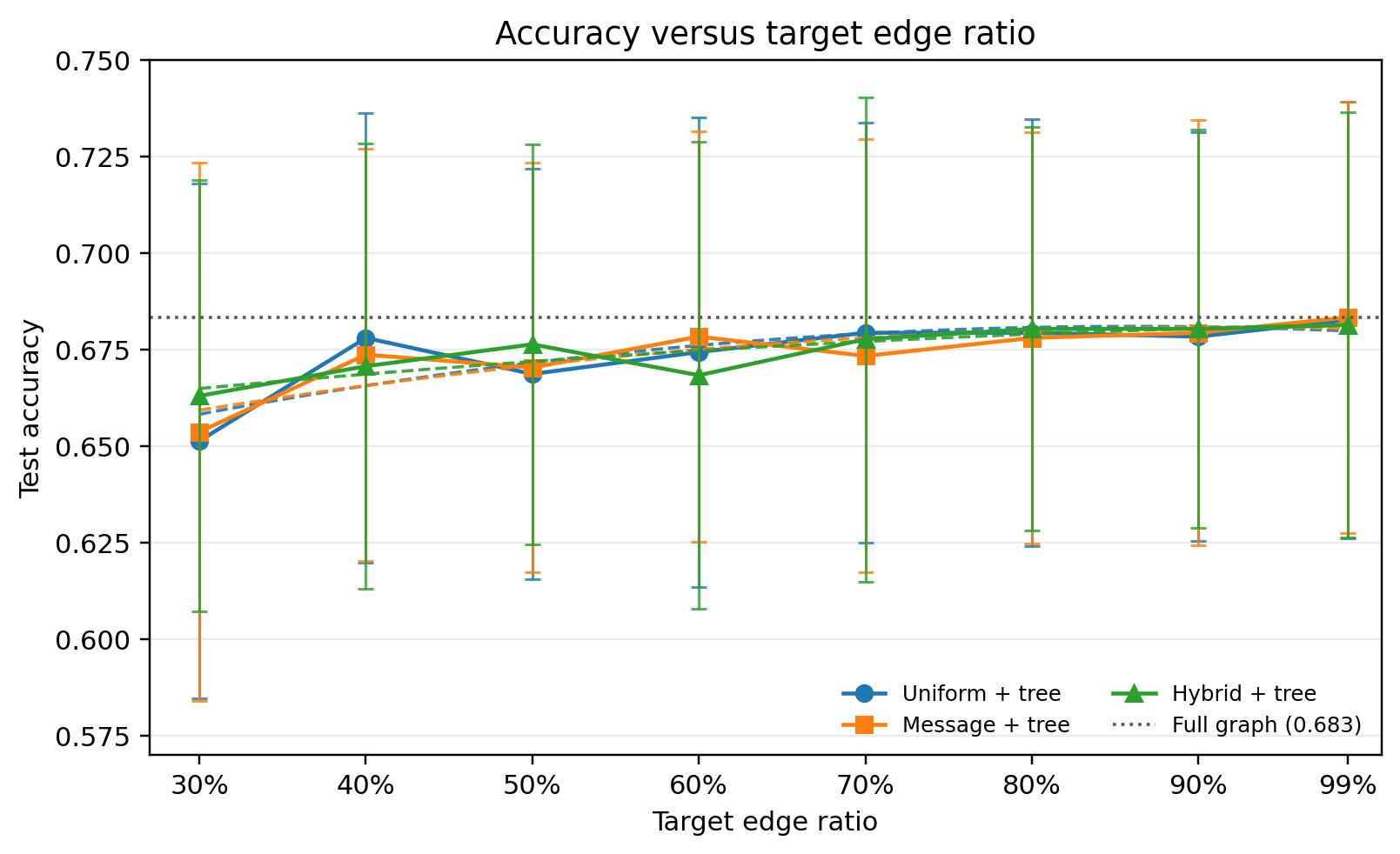}
\caption{Test accuracy versus target edge ratio. Markers show the mean at each of the eight tested ratios; error bars show one standard deviation. Dashed curves are quadratic least-squares fits using all eight means for each sampler and are descriptive only. The dotted horizontal line is the full-graph mean. }
\label{fig:accuracy-trend}
\end{figure}

Tables~\ref{tab:accuracy-results} and~\ref{tab:message-error-results} report test accuracy and relative message error, respectively. Figure~\ref{fig:accuracy-trend} plots accuracy against the target edge ratio. These results from the small synthetic pilot show consistent evidences to support the soundness of our sampling method and mathematical analysis. 

\begin{enumerate}
\item While sampling ratio grows from 30\% to 99\%, relative message errors in all three sampling methods fall consistently and significantly from about 1.2 to about 0.3. This illustrates
the message error and edge budget trade-off and supports our prioritized goal of preserving message importance. 
\item While sampling ratio grows from 30\% to 99\%, test accuracies in all three sampling methods remain in a narrow range with no loss but some minor gain. This supports out goal of not to sacrifice accuracy for message importance. 
\item In both the message score and the accuracy tests, performance of the proposed hybrid sampling lies in between of that of the message-preserving only sampling and uniform random sampling. This meets our design expectation as well as mathematical proofs. 
\end{enumerate}

These experimental results, in addition to the theoretic analysis, strongly support that the heuristic approach to preserve both structural features and message importance is a correct direction worth further exploration. However, honestly, most data from our modest pilot are not significant enough to compete with peer work. We believe that this on one hand is due to the limitation of the quality and scale of our synthetic data, while on the other hand due to the preliminary simple design of our proposal. 

\begin{table}[t]
\centering
\small
\caption{Test accuracy of node classification on the synthetic pilot.}
\label{tab:accuracy-results}
\begin{tabular}{lrrr}
\toprule
Target edge ratio & Uniform + tree & Message + tree & Hybrid + tree \\
\midrule
30\% & $0.651\pm0.067$ & $0.654\pm0.070$ & $0.663\pm0.056$ \\
40\% & $0.678\pm0.058$ & $0.674\pm0.053$ & $0.671\pm0.058$ \\
50\% & $0.669\pm0.053$ & $0.670\pm0.053$ & $0.676\pm0.052$ \\
60\% & $0.674\pm0.061$ & $0.678\pm0.053$ & $0.668\pm0.060$ \\
70\% & $0.679\pm0.054$ & $0.673\pm0.056$ & $0.678\pm0.063$ \\
80\% & $0.679\pm0.055$ & $0.678\pm0.053$ & $0.680\pm0.052$ \\
90\% & $0.678\pm0.053$ & $0.679\pm0.055$ & $0.680\pm0.051$ \\
95\% & $0.683\pm0.054$ & $0.681\pm0.055$ & $0.681\pm0.053$ \\
99\% & $0.683\pm0.056$ & $0.683\pm0.056$ & $0.681\pm0.055$ \\
\bottomrule
\end{tabular}
\end{table}

\begin{table}[t]
\centering
\small
\caption{Relative one-step message error $\|L_HX-LX\|_F/\|LX\|_F$ on the synthetic pilot.}
\label{tab:message-error-results}
\begin{tabular}{lrrr}
\toprule
Target edge ratio & Uniform + tree & Message + tree & Hybrid + tree \\
\midrule
30\% & $1.210\pm0.169$ & $1.216\pm0.173$ & $1.205\pm0.153$ \\
40\% & $0.716\pm0.046$ & $0.692\pm0.043$ & $0.711\pm0.049$ \\
50\% & $0.523\pm0.027$ & $0.507\pm0.023$ & $0.512\pm0.030$ \\
60\% & $0.397\pm0.017$ & $0.380\pm0.016$ & $0.387\pm0.018$ \\
70\% & $0.307\pm0.015$ & $0.290\pm0.011$ & $0.296\pm0.014$ \\
80\% & $0.228\pm0.011$ & $0.207\pm0.008$ & $0.218\pm0.013$ \\
90\% & $0.148\pm0.011$ & $0.127\pm0.007$ & $0.135\pm0.010$ \\
95\% & $0.099\pm0.009$ & $0.078\pm0.008$ & $0.086\pm0.008$ \\
99\% & $0.039\pm0.010$ & $0.027\pm0.004$ & $0.033\pm0.006$ \\
\bottomrule
\end{tabular}
\end{table}

\section{Conclusion and Future Work}
\label{sec:conclusion}

This paper studies edge sampling for approximating the one-step action
of the combinatorial Laplacian on a fixed input feature matrix. By
expressing the Laplacian message as a sum of edge-wise contributions,
we derive an unbiased estimator and its exact expected squared error.
With a fixed spanning-tree backbone, we characterize the minimum
message error achievable at each expected edge budget and the
corresponding error frontier. We also define a tree-based hybrid
sampler that combines feature-conditioned message scores with
effective-resistance leverage scores. An exact excess-error
decomposition quantifies the cost of this hybrid allocation relative
to the message-optimal allocation at the same budget. The controlled synthetic experiments provide preliminary empirical evidence consistent with the analysis. Together, these
results provide a sound approach to study how edge selection affects
one-step messages while retaining graph connectivity.

Future work will address these directions. First, the method should
be evaluated on larger public attributed-graph benchmarks and
integrated into standard GCN, GraphSAGE, and GAT pipelines. Such
evaluations should report predictive performance as well as graph
construction and propagation costs. Second, the analysis should be
extended beyond the one-step action of the combinatorial Laplacian on
fixed input features to normalized propagation operators and the
evolving hidden representations of multi-layer networks. Developing
guarantees over a feature-generated Krylov space is a related
direction. Finally, we shall further explore alternative sampling
strategies and conduct systematic experiments across graph sizes,
feature conditions, and mixture parameters, with uncertainty
estimates. 
\bibliographystyle{plainnat}
\bibliography{references_r1}

@article{eppstein1997sparsification,
  author = {Eppstein, David and Galil, Zvi and Italiano, Giuseppe F. and Nissenzweig, Amnon},
  title = {Sparsification---A Technique for Speeding Up Dynamic Graph Algorithms},
  journal = {Journal of the ACM},
  volume = {44},
  number = {5},
  pages = {669--696},
  year = {1997},
  doi = {10.1145/265910.265914}
}

@article{karger1999random,
  author = {Karger, David R.},
  title = {Random Sampling in Cut, Flow, and Network Design Problems},
  journal = {Mathematics of Operations Research},
  volume = {24},
  number = {2},
  pages = {383--413},
  year = {1999},
  doi = {10.1287/moor.24.2.383}
}

@article{spielman2011spectral,
  author = {Spielman, Daniel A. and Teng, Shang-Hua},
  title = {Spectral Sparsification of Graphs},
  journal = {SIAM Journal on Computing},
  volume = {40},
  number = {4},
  pages = {981--1025},
  year = {2011},
  doi = {10.1137/08074489X}
}

@article{spielman2011effective,
  author = {Spielman, Daniel A. and Srivastava, Nikhil},
  title = {Graph Sparsification by Effective Resistances},
  journal = {SIAM Journal on Computing},
  volume = {40},
  number = {6},
  pages = {1913--1926},
  year = {2011},
  doi = {10.1137/080734029}
}

@article{batson2012twice,
  author = {Batson, Joshua and Spielman, Daniel A. and Srivastava, Nikhil},
  title = {Twice-Ramanujan Sparsifiers},
  journal = {SIAM Journal on Computing},
  volume = {41},
  number = {6},
  pages = {1704--1721},
  year = {2012},
  doi = {10.1137/090772873}
}

@article{fung2019framework,
  author = {Fung, Wai Shing and Hariharan, Ramesh and Harvey, Nicholas J. A. and Panigrahi, Debmalya},
  title = {A General Framework for Graph Sparsification},
  journal = {SIAM Journal on Computing},
  volume = {48},
  number = {4},
  year = {2019},
  doi = {10.1137/16M1091666}
}

@inproceedings{kipf2017gcn,
  author = {Kipf, Thomas N. and Welling, Max},
  title = {Semi-Supervised Classification with Graph Convolutional Networks},
  booktitle = {International Conference on Learning Representations},
  year = {2017}
}

@inproceedings{hamilton2017graphsage,
  author = {Hamilton, William L. and Ying, Rex and Leskovec, Jure},
  title = {Inductive Representation Learning on Large Graphs},
  booktitle = {Advances in Neural Information Processing Systems},
  volume = {30},
  year = {2017}
}

@inproceedings{velickovic2018gat,
  author = {Veli{\v{c}}kovi{\'c}, Petar and Cucurull, Guillem and Casanova, Arantxa and Romero, Adriana and Li{\`o}, Pietro and Bengio, Yoshua},
  title = {Graph Attention Networks},
  booktitle = {International Conference on Learning Representations},
  year = {2018}
}

@inproceedings{chen2018fastgcn,
  author = {Chen, Jie and Ma, Tengfei and Xiao, Cao},
  title = {{FastGCN}: Fast Learning with Graph Convolutional Networks via Importance Sampling},
  booktitle = {International Conference on Learning Representations},
  year = {2018}
}

@inproceedings{huang2018adaptive,
  author = {Huang, Wenbing and Zhang, Tong and Rong, Yu and Huang, Junzhou},
  title = {Adaptive Sampling Towards Fast Graph Representation Learning},
  booktitle = {Advances in Neural Information Processing Systems},
  volume = {31},
  year = {2018}
}

@inproceedings{zou2019ladies,
  author = {Zou, Difan and Hu, Ziniu and Wang, Yewen and Jiang, Song and Sun, Yizhou and Gu, Quanquan},
  title = {Layer-Dependent Importance Sampling for Training Deep and Large Graph Convolutional Networks},
  booktitle = {Advances in Neural Information Processing Systems},
  volume = {32},
  year = {2019}
}

@inproceedings{chiang2019clustergcn,
  author = {Chiang, Wei-Lin and Liu, Xuanqing and Si, Si and Li, Yang and Bengio, Samy and Hsieh, Cho-Jui},
  title = {{Cluster-GCN}: An Efficient Algorithm for Training Deep and Large Graph Convolutional Networks},
  booktitle = {Proceedings of the 25th ACM SIGKDD International Conference on Knowledge Discovery \& Data Mining},
  pages = {257--266},
  year = {2019}
}

@inproceedings{wu2019sgc,
  author = {Wu, Felix and Souza, Amauri and Zhang, Tianyi and Fifty, Christopher and Yu, Tao and Weinberger, Kilian},
  title = {Simplifying Graph Convolutional Networks},
  booktitle = {Proceedings of the 36th International Conference on Machine Learning},
  series = {Proceedings of Machine Learning Research},
  volume = {97},
  pages = {6861--6871},
  year = {2019}
}

@inproceedings{zeng2020graphsaint,
  author = {Zeng, Hanqing and Zhou, Hongkuan and Srivastava, Ajitesh and Kannan, Rajgopal and Prasanna, Viktor},
  title = {{GraphSAINT}: Graph Sampling Based Inductive Learning Method},
  booktitle = {International Conference on Learning Representations},
  year = {2020}
}

@inproceedings{rong2020dropedge,
  author = {Rong, Yu and Huang, Wenbing and Xu, Tingyang and Huang, Junzhou},
  title = {{DropEdge}: Towards Deep Graph Convolutional Networks on Node Classification},
  booktitle = {International Conference on Learning Representations},
  year = {2020}
}

@article{srinivasa2020fastgat,
  author = {Srinivasa, Rakshith S. and Xiao, Cao and Glass, Lucas and Romberg, Justin and Sun, Jimeng},
  title = {Fast Graph Attention Networks Using Effective Resistance Based Graph Sparsification},
  journal = {arXiv preprint arXiv:2006.08796},
  year = {2020}
}

@article{shen2024gper,
  author = {Shen, Xu and Li{\`o}, Pietro and Yang, Lintao and Yuan, Ru and others},
  title = {Graph Rewiring and Preprocessing for Graph Neural Networks Based on Effective Resistance},
  journal = {IEEE Transactions on Knowledge and Data Engineering},
  volume = {36},
  number = {11},
  pages = {6330--6343},
  year = {2024},
  doi = {10.1109/TKDE.2024.3397692}
}

@inproceedings{hashemi2024survey,
  author = {Hashemi, Mohammad and Gong, Shengbo and Ni, Juntong and Fan, Wenqi and Prakash, B. Aditya and Jin, Wei},
  title = {A Comprehensive Survey on Graph Reduction: Sparsification, Coarsening, and Condensation},
  booktitle = {Proceedings of the Thirty-Third International Joint Conference on Artificial Intelligence},
  pages = {8058--8066},
  year = {2024},
  doi = {10.24963/ijcai.2024/891}
}

@article{chen2024demystifying,
  author = {Chen, Yuhan and Ye, Haojie and Vedula, Sanketh and Bronstein, Alex and Dreslinski, Ronald and Mudge, Trevor and Talati, Nishil},
  title = {Demystifying Graph Sparsification Algorithms in Graph Properties Preservation},
  journal = {Proceedings of the VLDB Endowment},
  year = {2024}
}

@inproceedings{vatter2024size,
  author = {Vatter, Jana and Rochau, Maurice L. and Mayer, Ruben and Jacobsen, Hans-Arno},
  title = {Size Does (Not) Matter? Sparsification and Graph Neural Network Sampling for Large-scale Graphs},
  booktitle = {Proceedings of the 3rd International Workshop on Large-Scale Graph Data Analytics (LSGDA), VLDB Workshops},
  year = {2024}
}

@article{ding2024coresets,
  author = {Ding, Mucong and He, Yinhan and Li, Jundong and Huang, Furong},
  title = {Spectral Greedy Coresets for Graph Neural Networks},
  journal = {arXiv preprint arXiv:2405.17404},
  year = {2024}
}

@inproceedings{zhang2025mog,
  author = {Zhang, Guibin and Sun, Xiangguo and Yue, Yanwei and Jiang, Chonghe and Wang, Kun and Chen, Tianlong and Pan, Shirui},
  title = {Graph Sparsification via Mixture of Graphs},
  booktitle = {International Conference on Learning Representations},
  year = {2025}
}

@article{das2025sgsgnn,
  author = {Das, Siddhartha Shankar and Arafat, Naheed Anjum and Rahman, Muftiqur and Ferdous, S. M. and Pothen, Alex and Halappanavar, Mahantesh M.},
  title = {{SGS-GNN}: A Supervised Graph Sparsification Method for Graph Neural Networks},
  journal = {arXiv preprint arXiv:2502.10208},
  year = {2025}
}

@inproceedings{ding2025spectral,
  author = {Ding, Haipeng and Wei, Zhewei and Ye, Yuhang},
  title = {Large-Scale Spectral Graph Neural Networks via Laplacian Sparsification},
  booktitle = {Proceedings of the 31st ACM SIGKDD Conference on Knowledge Discovery and Data Mining},
  pages = {213--223},
  year = {2025},
  doi = {10.1145/3690624.3709241}
}

@book{boyd2004convex,
  author    = {Boyd, Stephen and Vandenberghe, Lieven},
  title     = {Convex Optimization},
  publisher = {Cambridge University Press},
  year      = {2004},
  address   = {Cambridge}
}

\end{document}